\documentclass[11pt]{article}

\usepackage[final]{acl}

\usepackage{times}
\usepackage{latexsym}
\usepackage{amssymb}
\usepackage{amsmath}
\usepackage{hyperref}

\usepackage[T1]{fontenc}

\usepackage[utf8]{inputenc}

\usepackage{microtype}

\usepackage{inconsolata}

\usepackage{graphicx}
\usepackage{subcaption}

\def\@makefnmark{\hbox{\@textsuperscript{\normalfont\@thefnmark}}}
\makeatother

\newcommand{\starfootnotemark}{%
  \begingroup
    \renewcommand{\thefootnote}{\fnsymbol{footnote}}%
    \footnotemark[1]%
  \endgroup
}

\title{CobwebTM: Probabilistic Concept Formation for Lifelong and Hierarchical Topic Modeling}

\author{Karthik Singaravadivelan\thanks{ Equal Contribution}, Anant Gupta\starfootnotemark, Zekun Wang, Christopher J. MacLellan \\
  College of Computing \\
  Georgia Institute of Technology \\
  Atlanta, GA 30332 USA \\
  \{\texttt{ksingara3}, \texttt{agupta886}, \texttt{zwang910}\}@gatech.edu }

\begin{document}
\maketitle
\begin{abstract}
Topic modeling seeks to uncover latent semantic structure in text corpora with minimal supervision. Neural approaches achieve strong performance but require extensive tuning and struggle with lifelong learning due to catastrophic forgetting and fixed capacity, while classical probabilistic models lack flexibility and adaptability to streaming data.
We introduce \textsc{CobwebTM}, a low-parameter lifelong hierarchical topic model based on incremental probabilistic concept formation. By adapting the Cobweb algorithm to continuous document embeddings, \textsc{CobwebTM} constructs semantic hierarchies online, enabling unsupervised topic discovery, dynamic topic creation, and hierarchical organization without predefining the number of topics.
Across diverse datasets, \textsc{CobwebTM} achieves strong topic coherence, stable topics over time, and high-quality hierarchies, demonstrating that incremental symbolic concept formation combined with pretrained representations is an efficient approach to topic modeling.\let\thefootnote\relax
\footnotetext{Code available at \url{https://github.com/Teachable-AI-Lab/cobweb-language-embedding}}
\end{abstract}

\section{Introduction}
\begin{figure*}[ht]
\centering
\begin{subfigure}[b]{0.56\textwidth}
\centering
\includegraphics[width=\linewidth]{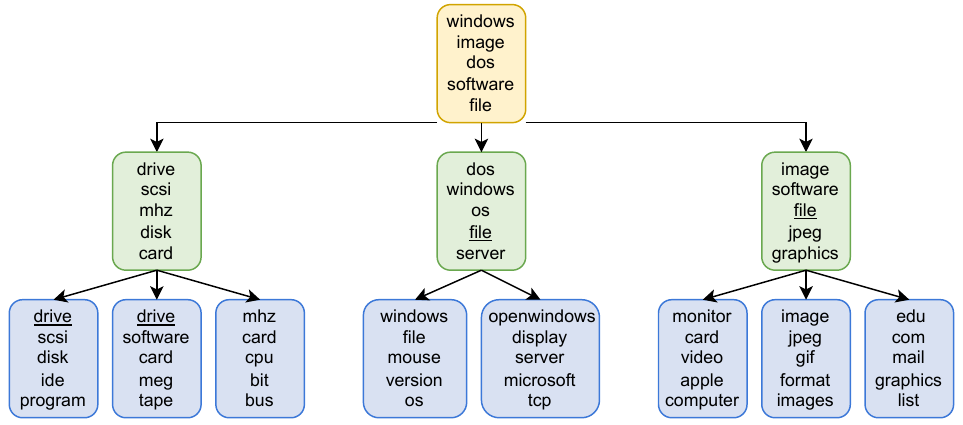}
\caption{Subtree from 20 Newsgroups dataset}
\label{fig:20ng_windows}
\end{subfigure}\hfill
\begin{subfigure}[b]{0.40\textwidth}
\centering
\includegraphics[width=\linewidth]{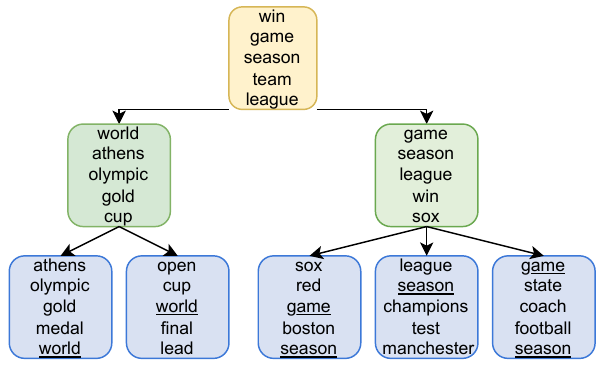}
\caption{Subtree from AG news dataset}
\label{fig:agnews_sports}
\end{subfigure}
\caption{A visualization of three levels of the hierarchy induced by \textsc{CobwebTM}. For each node, we display the top five representative words extracting using the c-tf-idf procedure described in Section~\ref{sec:ctfidf}. Words that appear in multiple nodes at the same level are underlined to highlight shared semantic content across sibling topics.}
\label{fig:hierarchy_viz}
\end{figure*}
Topic modeling seeks to uncover latent semantic structure in large document collections by grouping text into coherent topics. It is a fundamental tool for document organization, corpus exploration, and information retrieval, particularly in settings where labeled data is unavailable. As modern text corpora grow in scale, diversity, and temporal span, effective topic modeling increasingly requires methods that support unsupervised topic discovery, adapt to streaming data, and represent topics at multiple levels of abstraction.

Early work in topic modeling was dominated by probabilistic generative models, most notably Latent Dirichlet Allocation (LDA)~\citep{LDA-10.5555/944919.944937}. While influential, LDA requires the number of topics to be specified in advance, assumes independence between topics, and relies on bag-of-words representations that ignore semantic similarity between words. These assumptions limit its ability to model imbalanced, correlated, or evolving topics, making it poorly suited for lifelong or streaming settings.

Recent advances in representation learning have led to neural topic models that leverage dense document embeddings~\citep{zheng2013supervisedneuralautoregressivetopic, Wu_2024_survey}. These approaches often achieve improved topic coherence and richer semantic representations, but at the cost of increased complexity. Neural topic models are typically highly parameterized, sensitive to hyperparameter choices, and trained in batch settings that assume access to the full corpus. Consequently, they struggle in lifelong learning scenarios where data arrives incrementally and topic structure must evolve over time. Moreover, neural architectures are prone to catastrophic forgetting, causing previously learned topics to degrade as new data is introduced.

Lifelong topic modeling addresses these challenges by updating topics incrementally as new documents arrive. Methods such as Online LDA~\citep{OnlineLDA-NIPS2010_71f6278d} and neural lifelong topic models mitigate some scalability issues but retain key limitations, including fixed topic capacity, limited topic restructuring, and reliance on corpus-specific training. More recent embedding-based pipelines replace static clustering with incremental clustering algorithms, yet these methods remain sensitive to parameter choices and typically lack principled mechanisms for organizing topics at multiple levels of abstraction.

In practice, however, topic structure is inherently hierarchical: broad themes naturally decompose into progressively finer subtopics. Capturing such hierarchical organization improves interpretability and allows models to represent semantic relationships between topics rather than treating them as independent clusters. Consequently, hierarchical topic models have been widely explored in both probabilistic and neural frameworks~\citep{hldablei, koltcov2021analysistuninghierarchicaltopic}. These approaches aim to learn topic trees that capture varying levels of abstraction within a corpus.

Despite their promise, many hierarchical topic models rely on fixed-depth latent structures or require batch training over the full corpus, limiting their applicability in dynamic or streaming environments. In many modern systems, hierarchy is therefore imposed post hoc after flat topic discovery, rather than learned incrementally as the data evolves. This disconnect between lifelong learning and hierarchical structure motivates the need for topic modeling approaches that can simultaneously support incremental updates and flexible hierarchical organization.

In this work, we revisit incremental concept formation as an alternative paradigm for topic modeling. We introduce \textsc{CobwebTM}, a lifelong hierarchical topic modeling framework based on the Cobweb algorithm \citep{fisher1987cobweb} for probabilistic concept formation. By adapting Cobweb to operate over continuous document embeddings, \textsc{CobwebTM} incrementally constructs a semantic hierarchy as documents arrive, enabling unsupervised topic discovery without predefining the number of topics. 

Our contributions are threefold: (1) we introduce \textsc{CobwebTM}, an incremental hierarchical topic modeling framework for unsupervised topic discovery over streaming text; (2) we show that probabilistic concept formation in embedding space provides a simple yet effective mechanism for lifelong topic modeling without catastrophic forgetting or fixed topic capacity; and (3) through extensive empirical evaluation, we demonstrate that \textsc{CobwebTM} matches or outperforms recent neural and clustering-based methods in both topic quality and hierarchical structure.

\section{Related Works}
\subsection{Lifelong Topic Modeling}

Online LDA~\citep{OnlineLDA-NIPS2010_71f6278d} is the most widely used lifelong topic model, updating global topics via mini-batch variational inference. However, it inherits LDA’s bag-of-words assumption, requires a predefined number of topics, and lacks mechanisms for restructuring topics as new data arrives.

Most neural topic models are trained in batch settings and struggle with sequential updates without retraining~\citep{Wu_2024}. They are also prone to catastrophic forgetting~\citep{luo2025empiricalstudycatastrophicforgetting}. Mitigation techniques such as replay or elastic weight consolidation~\citep{lifelong-neuraltm-10.5555/3524938.3525304} reduce forgetting but still rely on fixed latent dimensions.

Embedding-based pipelines instead perform topic discovery through clustering over neural representations. BERTopic~\citep{grootendorst2022bertopic}, for example, combines transformer embeddings and clustering. Lifelong variants replace static clustering with incremental methods such as DBStream~\citep{DBSTREAM-6906426} or Mini-Batch KMeans~\citep{mini-batch-kmeans-10.1145/1772690.1772862}, though these approaches typically assume flat clustering and remain sensitive to parameter choices. Recent approaches such as TopicGPT~\citep{pham2024topicgpt} and FASTopic~\citep{wu2024fastopic} improve topic quality through LLM-based generation or embedding-level semantic modeling, but are either computationally expensive at scale or do not support hierarchical and incremental topic discovery.

\subsection{Hierarchical Topic Modeling}

Hierarchical topic models organize topics across levels of abstraction. Early Bayesian approaches such as hLDA~\cite{hldablei} and related models~\cite{mimno2007mixtures,perotte2011hierarchically} learn topic trees through generative processes. More recent methods construct hierarchies over embedding-based topic representations. Examples include CluHTM~\cite{viegas2020cluhtm}, HyHTM~\cite{shahid2023hyhtm}, and hierarchical variants of BERTopic~\cite{grootendorst2022bertopic}, which typically derive hierarchies through clustering or linkage procedures applied after flat topic discovery. Neural hierarchical topic models further learn structured latent representations using VAEs~\cite{kingma2013auto}, including tree-based~\cite{isonuma2020tree}, fixed-depth~\cite{duan2021sawetm}, and geometrically regularized models~\cite{wu2024traco, lu2024self}. However, these models are generally trained in batch settings and impose structural constraints that limit their flexibility in lifelong or streaming scenarios.

\subsection{Incremental Concept Formation}

Humans organize knowledge hierarchically using prototypes and graded category membership~\citep{rosch1975family}. Incremental clustering methods formalize this process by building taxonomies whose internal nodes summarize concept-level statistics.

Cobweb~\cite{fisher1987cobweb} incrementally constructs a probabilistic taxonomy through conceptual clustering, dynamically creating and restructuring nodes to maximize category utility. Recent work has extended Cobweb to neural settings and demonstrated robustness in vision and language tasks~\citep{efficientinductionlanguagemodels2022, convolutionalcobweb2021, wang2025taxonomic, barari2024incrementalconceptformationvisual, barari2024avoiding, LIAN2025101371}.

Unlike probabilistic topic models such as LDA, which directly learn $P(\text{word}|\text{topic})$ and $P(\text{topic}|\text{document})$ through Dirichlet priors, our approach derives these quantities through clustering in embedding space. Continuous Cobweb incrementally partitions transformer document embeddings into a hierarchical mixture of clusters, estimating document–topic associations via category utility. Topic–word distributions are computed post hoc using class-based TF–IDF over the documents assigned to each node.

\section{Methodology}

We propose \textsc{CobwebTM}, a topic modeling framework that incrementally organizes document embeddings into a dynamic semantic hierarchy. Unlike batch clustering methods such as k-Means or HDBSCAN, \textsc{CobwebTM} supports continual updates without retraining through a two-step neuro-symbolic process.

First, we perform document–topic inference directly in the latent space of pretrained transformer embeddings. Assuming the embedding space reflects an underlying mixture of topics, we apply the continuous Cobweb algorithm to incrementally partition the space, assigning each document to a node that maximizes category utility. This procedure produces a hierarchical clustering that implicitly defines the document–topic distribution.

Second, we derive topic–word representations from the resulting hierarchy. Each node represents a topic defined by the documents in its subtree. Treating nodes as classes, we compute word–topic distributions using c-TF-IDF, producing interpretable topic descriptors from the highest-ranked words.

\subsection{Probabilistic Concept Formation}

At the core of our approach is a variant of Cobweb adapted for continuous-valued attributes \citep{barari2024incrementalconceptformationvisual}. Each concept node $c$ maintains a $D$-dimensional multivariate Gaussian with diagonal covariance,
\[
p(x \mid c) = \mathcal{N}\big(x;,\mu_c,,\mathrm{diag}(\sigma_c^2)\big),
\]
where $\mu_c \in \mathbb{R}^D$ is the node mean and $\sigma_c^2 \in \mathbb{R}^D$ is the variance vector. These statistics are updated incrementally as new documents are incorporated.

Cobweb constructs a hierarchy of concepts online. Given a new document embedding $x$, the algorithm performs a top-down search over the tree guided by \textit{Category Utility} (CU) \citep{gluck1985information,cu-corter-gluck-1992-18581-001}. Following \citet{barari2024incrementalconceptformationvisual}, we adopt an information-theoretic formulation that measures the expected reduction in feature uncertainty obtained by knowing the child concept.

Let a parent node $c_p$ have children $\mathcal{C}(c_p)$, each with count $N_c$. The empirical probability of concept $c$ under the parent is
\begin{equation}
P(c \mid c_p) = \frac{N_c}{\sum_{c' \in \mathcal{C}(c_p)} N_{c'}} = \frac{N_c}{N_{c_p}}.
\end{equation}

We measure node uncertainty using the differential entropy of the Gaussian:
\begin{equation}
U(c) = \frac{1}{2}\sum_{d=1}^{D}\log\!\big(2\pi e,\sigma^{2}_{c,d}\big).
\end{equation}

The category utility of a parent node is then
\begin{equation}
\mathrm{CU}(c_p) = \sum_{c \in \mathcal{C}(c_p)} P(c \mid c_p)\,\bigl[ U(c_p) - U(c) \bigr].
\end{equation}

Maximizing CU favors partitions that reduce feature uncertainty while maintaining sufficient support, balancing intra-cluster similarity and inter-cluster separation. For continuous attributes, this corresponds to maximizing variance reduction induced by the partition, allowing Cobweb to determine the depth and breadth of the hierarchy without specifying the number of topics $K$.

At each node, Cobweb evaluates four operators to determine how $x$ should be incorporated into the hierarchy:
(1) insert $x$ into the best-matching existing child and update its Gaussian parameters;
(2) create a new singleton child node for $x$;
(3) merge the two best-matching children and assign $x$ to the merged node; or
(4) split the best-matching child, promoting its children to the current level.

\subsection{Topic Extraction}

\subsubsection{Hierarchical Topic Extraction}
\label{sec:ctfidf}
The Cobweb tree forms a multi-level topic hierarchy: the root represents the entire corpus, intermediate nodes correspond to increasingly specific topics, and leaves represent individual documents or fine-grained micro-topics.

To extract interpretable topics, we treat each node as a candidate topic and compute class-based TF–IDF (c-TF-IDF) scores following \citet{grootendorst2022bertopic}. For a node $C$, all documents in its subtree are concatenated into a single document. The importance of word $w$ in topic $C$ is then computed as
\begin{equation}
W_{C,w} = tf_{C,w} \times \log \left(1 + \frac{A}{f_w} \right),
\end{equation}
where $tf_{C,w}$ is the frequency of word $w$ within topic $C$, $f_w$ is the frequency of $w$ across all topics, and $A$ is the average number of words per topic. The highest-scoring words serve as descriptive keywords for the topic.

\subsubsection{Dynamic Flat Topic Extraction}

While the hierarchical structure provides rich semantic organization, many incremental topic modeling baselines operate on a flat set of topics. To enable direct comparison and support applications requiring fixed topic sets, we extract a dynamic flat partition from the hierarchy.

Because Cobweb is incremental, documents may appear at varying depths depending on their semantic specificity. To obtain a coherent flat topic set, we identify a cut through the tree that balances topic coverage and granularity. Specifically, we traverse the hierarchy top-down and select nodes such that (1) the number of selected nodes does not exceed a user-defined \texttt{max\_clusters}, and (2) the ratio of leaf nodes to total nodes in the layer does not exceed a \texttt{leaf\_ratio} threshold.

This procedure filters out shallow outliers near the root while grouping deeper semantically similar documents into stable clusters. As new documents arrive and the hierarchy evolves, the cut is recomputed, allowing the flat topic representation to adapt over time. Consequently, \textsc{CobwebTM} supports both a flat topic model for benchmarking and a full hierarchical representation that enables exploration across levels of abstraction.

\section{Lifelong Topic Modeling}
\label{sec:lifelongtm}

We first evaluate the performance of \textsc{CobwebTM} in a lifelong topic modeling setting. The primary objective is to assess the model's ability to maintain coherent topics, ensure stability across time steps, and adapt to new data without catastrophic forgetting or the need for extensive retraining.

\subsection{Experimental Setup}
\label{sec:lfltm_setup}

\paragraph{Datasets.}
We utilize three datasets suited for temporal analysis: the \textbf{Spatiotemporal News Dataset}~\citep{spatiotemporalDataset2025}, the \textbf{Stack Overflow Dataset}~\citep{STACKOVERFLOW}, and the \textbf{TweetNER7 Dataset}~\citep{ushio-etal-2022-tweet}. These datasets represent varying document lengths and stream velocities. The Spatiotemporal News and TweetNER7 datasets are temporally ordered to reflect real-world streams with intermittent topics, while the Stack Overflow dataset is randomly shuffled to simulate an approximately uniform incremental topic distribution. Detailed preprocessing steps and dataset statistics are provided in Appendix~\ref{app:datasets}.

\paragraph{Baselines.}
We compare \textsc{CobwebTM} against a range of incremental and static baselines. First, we evaluate \textbf{Online LDA} \citep{OnlineLDA-NIPS2010_71f6278d}, an incremental variational Bayes variant of Latent Dirichlet Allocation. Second, we include \textbf{Lifelong NTM} \citep{lifelong-neuraltm-10.5555/3524938.3525304}, a neural topic model extending DocNADE with Elastic Weight Consolidation and experience replay. Third, we employ \textbf{BERTopic (Incremental)} pipelines using incremental clustering algorithms, specifically \textbf{DBSTREAM} and \textbf{MiniBatchKMeans}. Finally, to benchmark against non-incremental upper bounds, we evaluate \textbf{BERTopic (Re-fit)} pipelines that re-train \textbf{HDBSCAN} and \textbf{KMeans} from scratch on the accumulated corpus at each time step.

\paragraph{Implementation Details.}
We use the embedding model RoBERTa Large (dimension size of 1,024) \citep{liu2019robertarobustlyoptimizedbert} for all pipelines which require it to ensure consistent feature spaces. We vary the initial batch size (default 2,000 documents, sensitivity analysis at 500). Unlike neural baselines, \textsc{CobwebTM} does not require large batch sizes for stability, so we fix the successive batch sizes at a middle ground of 125 documents. Topics between consecutive batches are matched using a greedy alignment strategy based on cosine similarity of topic embeddings.

\begin{table*}[ht]
\centering
\begin{tabular}{|l|ccccc|}
\hline
Method & ${C_v}_f$ & $\Delta C_v$\% & ARI & TCD & ISIM$_f$ \\
\hline
CobwebTM (ours)  & \textbf{0.741} & \textbf{110.82} & 0.915 & \textbf{0.000} & \textbf{0.191} \\
DBSTREAM          & 0.602 & \underline{71.55} & 0.594 & 0.202 & 0.200 \\
LifelongDocNADE   & 0.281 & $-11.61$ & 0.479 & 0.369 & \underline{0.194} \\
MiniBatchKMeans   & 0.413 & $-0.89$ & \textbf{0.952} & 0.236 & 0.215 \\
OnlineLDA         & 0.317 & 7.06 & \underline{0.948} & \underline{0.033} & 0.216 \\
Refit-HDBSCAN     & \underline{0.673} & 7.35 & 0.944 & 0.053 & 0.196 \\
Refit-KMeans      & 0.369 & $-6.49$ & 0.549 & 0.120 & 0.224 \\
\hline
\end{tabular}
\caption{Lifelong topic modeling results on \textbf{TweetNER}. Best in bold, second-best underlined.}
\label{tab:tweetner}
\end{table*}

\begin{table*}[ht]
\centering
\begin{tabular}{|l|ccccc|}
\hline
Method & ${C_v}_f$ & $\Delta C_v$\% & ARI & TCD & ISIM$_f$ \\
\hline
CobwebTM (ours)  & \textbf{0.613} & \textbf{26.09} & \textbf{0.997} & \textbf{0.000} & 0.207 \\
DBSTREAM          & 0.457 & $-3.67$ & 0.012 & 0.155 & 0.188 \\
LifelongDocNADE   & 0.184 & $-75.07$ & 0.122 & 0.684 & 0.206 \\
MiniBatchKMeans   & 0.423 & $-9.67$ & \underline{0.955} & 0.245 & \underline{0.176} \\
OnlineLDA         & 0.520 & 10.71 & 0.897 & \underline{0.029} & 0.200 \\
Refit-HDBSCAN     & 0.425 & 1.81 & 0.735 & 0.050 & \textbf{0.157} \\
Refit-KMeans      & \underline{0.551} & \underline{15.98} & 0.416 & 0.205 & 0.224 \\
\hline
\end{tabular}
\caption{Lifelong topic modeling results on \textbf{Stack Overflow} Dataset. Best in bold, second-best underlined.}
\label{tab:stackexchange}
\end{table*}

\begin{table*}[ht]
\centering
\begin{tabular}{|l|ccccc|}
\hline
Method & ${C_v}_f$ & $\Delta C_v$\% & ARI & TCD & ISIM$_f$ \\
\hline
CobwebTM (ours)  & \textbf{0.796} & \textbf{57.51} & \textbf{0.984} & \textbf{0.000} & 0.208 \\
DBSTREAM          & 0.418 & $-10.29$ & 0.408 & 0.193 & \textbf{0.191} \\
LifelongDocNADE   & 0.238 & $-49.49$ & 0.127 & 0.462 & 0.196 \\
MiniBatchKMeans   & 0.646 & \underline{40.37} & \underline{0.962} & 0.203 & 0.196 \\
OnlineLDA         & 0.422 & 0.99 & 0.882 & \underline{0.053} & 0.206 \\
Refit-HDBSCAN     & \underline{0.657} & 19.11 & 0.891 & 0.067 & \underline{0.193} \\
Refit-KMeans      & 0.614 & 29.11 & 0.326 & 0.229 & 0.221 \\
\hline
\end{tabular}
\caption{Lifelong topic modeling results on \textbf{Spatiotemporal News} Dataset. Best in bold, second-best underlined.}
\label{tab:spatiotemporal}
\end{table*}
\begin{figure*}[ht]
    \includegraphics[width=\textwidth]{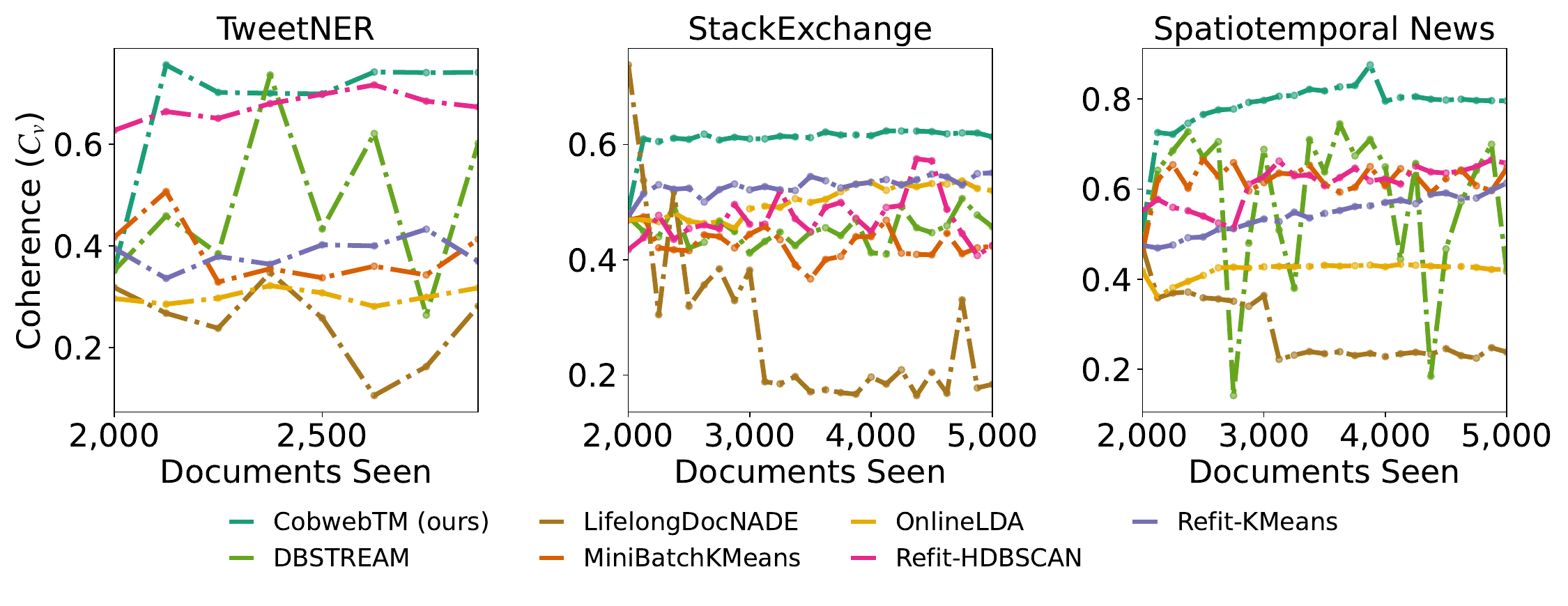}
    \caption{$C_v$ comparison on the StackOverflow, Spatiotemporal News, and TweetNER dataset.}
    \label{fig:incremental-cv-results}
\end{figure*}

\begin{figure*}[ht]
    \includegraphics[width=\textwidth]{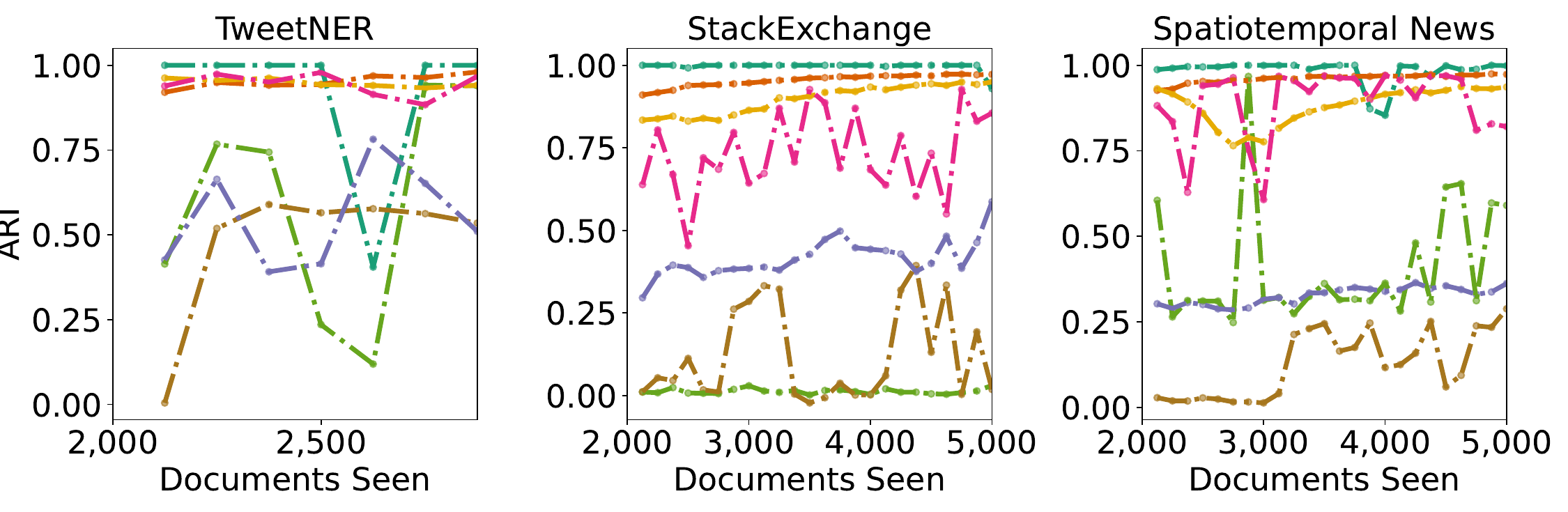}
    \caption{ARI comparison on the StackOverflow, Spatiotemporal News, and TweetNER dataset.}
    \label{fig:incremental-ari-results}
\end{figure*}

\begin{figure*}[ht]
        \includegraphics[width=\textwidth]{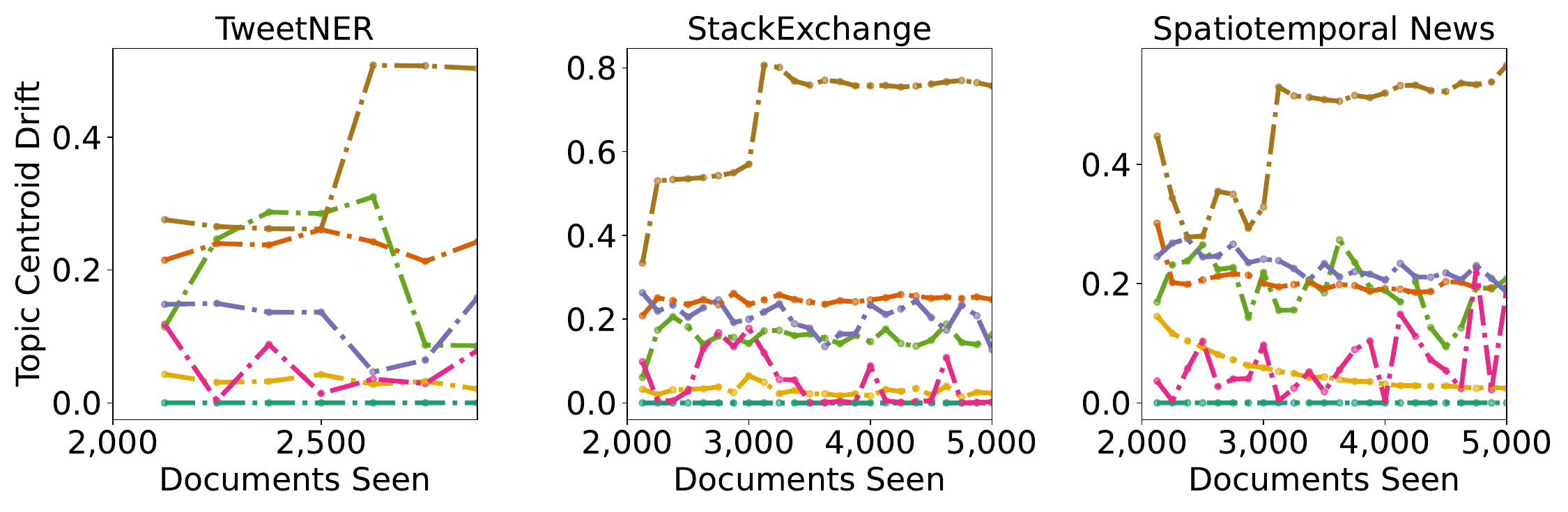}
    \caption{TCD comparison on the StackOverflow, Spatiotemporal News, and TweetNER dataset.}
    \label{fig:incremental-tcd-results}
\end{figure*}

\begin{figure*}[ht]
        \includegraphics[width=\textwidth]{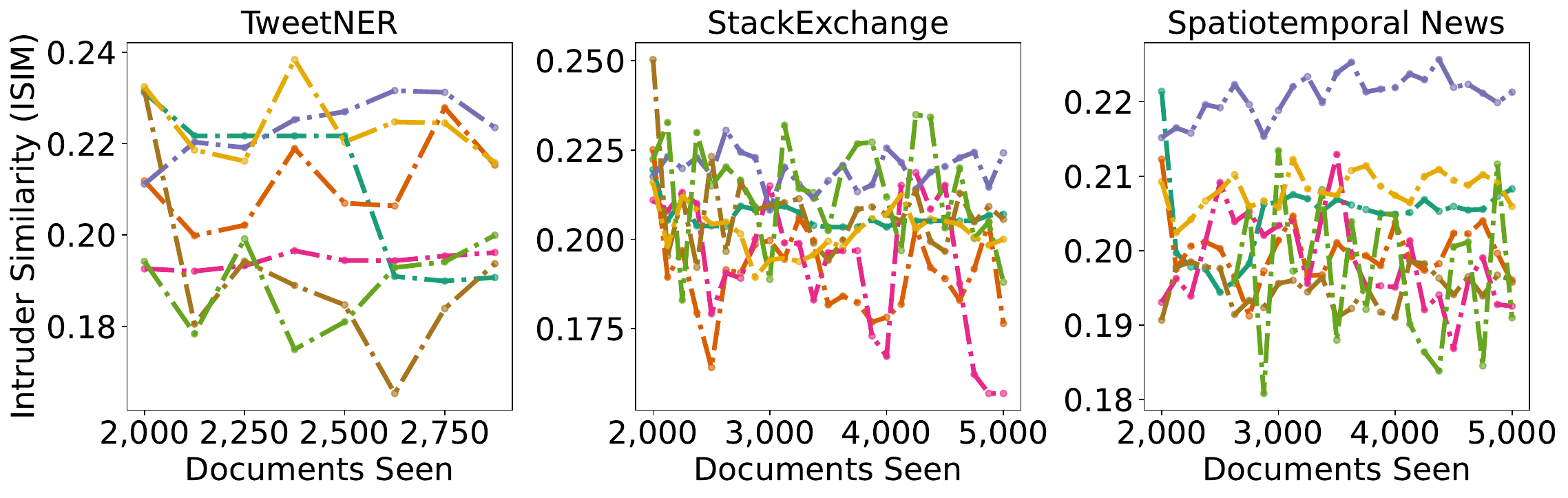}
    \caption{Intruder Similarity (ISIM) comparison on the StackOverflow, Spatiotemporal News, and TweetNER dataset.}
    \label{fig:incremental-isim-results}
\end{figure*}

\paragraph{Evaluation Metrics.}
We evaluate topic quality using the \textbf{Topic Coherence ($C_v$)} measure \cite{coherence-10.1145/2684822.2685324}, an indirect confirmation metric that combines an NPMI-style word co-occurrence statistic with a context-based similarity score computed over sliding windows. 
\[
C_v
= \frac{2}{N(N-1)}
\sum_{i<j}
\cos\!\left(
\vec{v}(w_i), \vec{v}(w_j)
\right)
\] 
where
$\vec{v}(w_i) = \{\mathrm{NPMI}(w_i, w_k)\}_{k=1}^N$.

Unlike purely count-based coherence measures, $C_v$ incorporates distributional information from the reference corpus, yielding scores that are comparable across batches. Coherence is computed over all data observed up to the current batch. 
We measure topic stability across consecutive batches using the \textbf{Adjusted Rand Index (ARI)} \citep{ARI-Topic-10.1007/978-3-662-44848-9_32}, computed between document–topic assignments from the previous and current batches. Higher ARI values indicate that topic assignments remain consistent over time, reflecting stable topic structure under incremental updates.
To quantify semantic drift of topics across batches, we compute \textbf{Topic Centroid Drift (TCD)} for matched topics. For each topic, we represent its semantic centroid using topic embeddings and define drift as one minus the cosine similarity between the current and previous centroids. The reported score is the mean drift across all matched topics. Values close to zero indicate minimal semantic change and stable topics.
Lastly, to ensure that we achieve topics that are interpretable, we calculate \textbf{Intruder Similarity Score (ISIM)}. For each topic, we insert a random word and calculate the average cosine similarity between a set of topic words and the intruder word, with lower averages corresponding to tighter and more unique topics. 

We average results across three trials for each dataset on each configuration we report for \textsc{CobwebTM}. We set $\texttt{leaf\_ratio} = 0.15$ to adapt natural outlier pruning, and $\texttt{max\_clusters} = 1.3 \cdot K$, where $K$ is the recommended number of clusters for each dataset as provided by the original paper.

\subsection{Results}

As shown in Figures \ref{fig:incremental-cv-results}, \ref{fig:incremental-ari-results}, \ref{fig:incremental-tcd-results}, and \ref{fig:incremental-isim-results}, and in Tables \ref{tab:tweetner}, \ref{tab:stackexchange}, and \ref{tab:spatiotemporal} \textsc{CobwebTM} outperforms or performs competitively with baselines in $C_v$, ARI, TCD, and ISIM across all three datasets. We analyze specific results below.

\paragraph{Comparison to BERTopic.} Our comparisons to BERTopic pipelines boil down to the difference of the clustering algorithms. We find that DBSTREAM is extremely volatile on semantically dense datasets, resulting in poor coherence throughout batches. While BERTopic with MiniBatchKMeans initially outperforms \textsc{CobwebTM} in the TweetNER benchmark, \textsc{CobwebTM} shows more growth across all three datasets and eventually surpasses all methods in final performance, shown in Tables \ref{tab:tweetner}, \ref{tab:stackexchange}, and \ref{tab:spatiotemporal}. Additionally, \textsc{Cobweb} has no learning parameters, with the only two parameters being user-specified for granularity decisions in flat topic modeling. 

\paragraph{Comparison to Neural Methods.} The neural baseline exhibits degraded performance on the datasets, likely due to memory constraints and the datasets' large technical vocabulary. Even after pruning to the most frequent words, Lifelong NTM struggled to disambiguate topics beyond the initial batch as vocabulary diversity increased, as shown in Figure~\ref{fig:incremental-cv-results} and Figure~\ref{fig:incremental-tcd-results}. This behavior reflects a broader limitation of neural topic models, which must learn word semantics from the corpus itself rather than leveraging pretrained embeddings. Our approach, which combines pretrained encoder representations with symbolic learning, enables more stable incremental performance and reduced susceptibility to catastrophic forgetting.

\paragraph{Metrics Over Time.} \textsc{CobwebTM} shows strong increases in performance by coherence unlike other methods, as shown in Tables \ref{tab:tweetner}, \ref{tab:stackexchange}, and \ref{tab:spatiotemporal} and in Figures \ref{fig:incremental-ari-results} and \ref{fig:incremental-tcd-results}, indicating that incremental aggregation of documents does not hinder topic construction and maintenance. Additionally, \textsc{CobwebTM} has strong TCD and ARI between batches across all datasets, highlighting a quality of clustering stability and the ability to create new topics when needed without harming the sanctity of topics from previous batches.

\paragraph{Strong Temporal Clustering Stability.} The Adjusted Rand Index (ARI) \citep{ARI-Topic-10.1007/978-3-662-44848-9_32} measures the consistency of topic assignments across batches, reflecting a model’s resistance to topic drift and catastrophic forgetting. High ARI indicates that learned topics remain stable as new data arrive, while low ARI signals uncontrolled reorganization. As shown in Tables \ref{tab:tweetner}, \ref{tab:stackexchange}, and \ref{tab:spatiotemporal}, \textsc{CobwebTM} achieves consistently high ARI across datasets, with near-perfect scores on Stack Overflow (0.997) and Spatiotemporal News (0.984), demonstrating strong stability under incremental updates. Although MiniBatchKMeans scores slightly higher on TweetNER, its stability likely stems from fixed cluster constraints rather than adaptive topic evolution.

\paragraph{Robustness to Human-Centered Metrics.} Our results, shown in Figure \ref{fig:incremental-isim-results}, are shown for 5 topic-words, averaged across 15 intruder words. CobwebTM achieves competitive scores to other models, though ranges of 0.02 to 0.06 across datasets suggests the metric is largely encoder-dependent at this level of comparison. We hypothesize that the tight margins are as a result of the semantically dense content in the datasets we leverage for testing incremental topic modeling, and these preliminary studies support that CobwebTM provides human-interpretable topics competitive with other topic models.

\section{Hierarchical Topic Modeling}

In the second set of experiments, we evaluate the quality of the hierarchical structures generated by \textsc{CobwebTM}. We benchmark against state-of-the-art hierarchical topic models to verify that our incremental construction yields meaningful taxonomies.

\subsection{Experimental Setup}

\paragraph{Datasets.}
We use three standard benchmarks: \textbf{20 Newsgroups}~\cite{lang1995newsweeder}, \textbf{AG News}~\cite{zhang2015character}, and \textbf{Stack Overflow}~\cite{STACKOVERFLOW}.

\begin{table*}[ht]
\centering
\begin{tabular}{|l|ccc|ccc|ccc|}
\hline
Dataset & \multicolumn{3}{c|}{20 News Groups} & \multicolumn{3}{c|}{AG News} & \multicolumn{3}{c|}{Stack Exchange} \\
\cline{2-4}\cline{5-7}\cline{8-10}
Method & NPMI & PCC & SD & NPMI & PCC & SD & NPMI & PCC & SD \\
\hline
BERTopic-HDBSCAN & 0.161 & 0.045 & 0.600 & \underline{0.101} & \textbf{0.031} & 0.913 & \underline{0.130} & {0.036} & 0.891 \\
BERTopic-KMeans  & \underline{0.169} & \underline{0.091} & 0.875 & 0.067 & 0.005 & 0.856 & 0.125 & \underline{0.041} & 0.940 \\
TraCo              & 0.155 & -0.038 & 0.941 & -0.107 & -0.043 & 0.862 & -0.040 & -0.156 & 0.913 \\
BoxTM              & 0.047 & 0.055 & \textbf{0.993} & 0.002 & 0.005 & \textbf{0.998} & 0.019 & 0.022 & \textbf{0.996} \\
CobwebTM (ours)    & \textbf{0.206} & \textbf{0.141} & \underline{0.958} & \textbf{0.108} & \underline{0.027} & \underline{0.942} & \textbf{0.131} & \textbf{0.073} & \underline{0.959} \\
\hline
\end{tabular}
\caption{Comparison of hierarchical topic-modeling metrics across datasets.}
\label{tab:hierarchical_results}
\end{table*}

\paragraph{Baselines.}
We compare against three primary baselines: \textbf{TraCo}~\cite{wu2024traco}, a neural model using Optimal Transport for topic regularization; \textbf{BoxTM}~\cite{lu2024self}, a geometric approach modeling topics as hyper-rectangles, and \textbf{BERTopic (Hierarchical)}~\cite{grootendorst2022bertopic}, which uses agglomerative clustering on top of flat topics derived from HDBSCAN and KMeans.

\paragraph{Implementation Details.}
All embedding-based models use the same architectures and embedding dimensions as in Section~\ref{sec:lfltm_setup}. For the remaining models, we use the hyperparameters specified in their official implementations. Datasets for \textbf{BoxTM} and \textbf{TraCo} are preprocessed as described in Appendix~\ref{app:preprocessing}.

\paragraph{Evaluation Metrics.}
We assess the hierarchy using three metrics. \textbf{Topic Coherence (NPMI)} measures the semantic interpretability of individual topics using Normalized Pointwise Mutual Information~\cite{isonuma2020tree}. For word pairs within a topic, NPMI is defined as
\[
\mathrm{NPMI}(w_i, w_j) = \frac{\log \frac{P(w_i, w_j)}{P(w_i)P(w_j)}}{-\log P(w_i, w_j)}.
\]
We report the average NPMI score aggregated per level of the hierarchy.

\textbf{Parent-Child Coherence (PCC)} evaluates vertical semantic consistency by computing Cross-Level NPMI~\cite{chen2021evaluating} between each child topic $t$ and its parent $\pi(t)$. This metric averages NPMI scores over word pairs $(w_c, w_p)$ with $w_c \in W_t$ and $w_p \in W_{\pi(t)}$, excluding overlapping words to ensure that coherence reflects meaningful specialization rather than redundancy.

Finally, \textbf{Sibling Topic Diversity (SD)} assesses horizontal distinctiveness among sibling topics using an adaptation of Topic Diversity~\cite{dieng2020topic}. For a set of sibling topics $\mathcal{S}(p)$ under a common parent $p$, SD is computed as the ratio of words appearing in exactly one sibling topic to the total number of unique words across all siblings.

\subsection{Quantitative Results}

Table~\ref{tab:hierarchical_results} showcases quantitative comparisons across the three datasets using topic coherence (NPMI), parent--child coherence (PCC), and sibling diversity (SD). 

\paragraph{Topic Coherence.}
\textsc{CobwebTM} attains the highest NPMI on all three datasets, achieving 0.206 on 20~News Groups, 0.108 on AG~News, and 0.131 on Stack Exchange as per Table \ref{tab:hierarchical_results}. These results demonstrate that incremental, structure-aware hierarchy construction does not compromise topic interpretability. In contrast, BERTopic variants exhibit competitive but less consistent coherence, while TraCo and BoxTM suffer from substantially weaker topic quality, particularly on AG~News and Stack Exchange.

\paragraph{Vertical Consistency.}
Parent--child coherence further highlights the advantages of \textsc{CobwebTM}. As shown in Table~\ref{tab:hierarchical_results}, \textsc{CobwebTM} achieves the highest PCC across all datasets, indicating that its child topics reliably specialize their parents. Post-hoc hierarchical approaches such as BERTopic display mixed behavior, while TraCo frequently produces negative PCC values, suggesting poor semantic alignment between hierarchical levels.

\paragraph{Sibling Diversity.}
While BoxTM achieves near-perfect sibling diversity, its low NPMI and modest PCC indicate that this separation often comes at the expense of semantic coherence, reflecting over-segmentation. In contrast, \textsc{CobwebTM} maintains high sibling diversity (SD $\geq$ 0.94 across datasets) while simultaneously preserving strong topic coherence and vertical consistency. This balance suggests that \textsc{CobwebTM} induces meaningful distinctions among sibling topics without fragmenting semantic structure.

Taken together, our results demonstrate that \textsc{CobwebTM} offers a more holistic solution to hierarchical topic modeling, effectively balancing interpretability, hierarchical alignment, and structural diversity. These findings underscore the promise of incremental, structure-aware learning for inducing high-quality topic hierarchies.

\subsection{Qualitative Results}
In this section, we visualize the hierarchies created by \textsc{CobwebTM}. Figure \ref{fig:hierarchy_viz} shows sample hierarchies of topic summaries from the 20 Newsgroups and AG News datasets. The hierarchy captures meaningful semantic structure: parent nodes represent broad themes while children specialize into increasingly specific topics.

In the 20 Newsgroups hierarchy (Figure \ref{fig:20ng_windows}), a technology-focused root topic is partitioned into coherent subtopics related to memory specifications, operating systems, and graphical components. Importantly, \textsc{CobwebTM} is able to disentangle semantically similar terms that often cause confusion in embedding-based clustering. For example, within subtrees related to \textit{windows–dos–drive}, lower-level topics remain focused on computing concepts rather than drifting toward unrelated senses such as automobile-related uses of “drive.”

This behavior contrasts with hierarchies produced by common embedding-clustering pipelines such as BERTopic (using KMeans or HDBSCAN), where nearest-neighbor clustering can conflate semantically distinct senses of ambiguous terms. Because such methods rely primarily on local similarity, they implicitly treat all embedding dimensions as uniformly important, which can lead to mixed topic substructures.

Neural hierarchical topic models such as BoxTM and TraCo exhibit different behavior: while their hierarchies often contain lexically diverse terms, we observe weaker semantic coherence within some nodes. For instance, unrelated words like "butcher" and "boulder" may appear grouped together despite lacking a clear conceptual relationship.

In contrast, \textsc{CobwebTM} produces hierarchies that maintain stronger semantic consistency while separating ambiguous concepts. We attribute this behavior to Cobweb’s probabilistic formulation: each node maintains variance estimates over embedding dimensions, effectively weighting feature importance when forming clusters. By maximizing category utility, \textsc{CobwebTM} favors partitions that reduce uncertainty while preserving coherent semantic structure across levels of the hierarchy.

\section{Conclusion}
In this paper, we introduced \textsc{CobwebTM}, a lifelong hierarchical topic model that incrementally constructs probabilistic concept hierarchies from streaming text. By leveraging pretrained language model embeddings, \textsc{CobwebTM} exploits the geometric structure of the embedding space to induce semantically coherent topic hierarchies without requiring task-specific hyperparameter tuning. The proposed framework naturally supports lifelong learning, allowing the hierarchy to evolve as new documents arrive. Extensive experiments across multiple hierarchical and lifelong topic modeling benchmarks demonstrate that \textsc{CobwebTM} consistently outperforms recent methods in both topic quality and hierarchical organization, highlighting its effectiveness as a scalable and adaptive topic modeling approach.

\section*{Limitations}
While \textsc{CobwebTM} demonstrates strong empirical performance, several limitations remain. First, although the model incrementally induces hierarchical clusters in embedding space, topic word extraction relies on post hoc aggregation of documents at each node, rather than being directly generated during hierarchy construction. Second, \textsc{CobwebTM} depends on pretrained document embeddings; consequently, the quality of the learned hierarchy is constrained by the representational capacity of the underlying encoder, and semantic distinctions poorly captured in the embedding space may be lost. Third, the incremental clustering process is sensitive to document arrival order, particularly in non-stationary streams. Although local restructuring operations mitigate this effect, globally optimal hierarchies are not guaranteed. Finally, while well suited to lifelong learning, maintaining statistics over large hierarchies incurs growing memory and computational costs, potentially necessitating pruning, compression, or partitioning strategies in long-running deployments.
\paragraph{Future Work.}
A natural extension of \textsc{CobwebTM} is multimodal topic modeling. Because the framework operates on continuous representations, images and audio can be incorporated via contrastively trained vision–language models and text-based encoders, enabling heterogeneous data to be organized within a shared hierarchical topic space without modifying the underlying algorithm. Additionally, leveraging node-level statistics to generate topic summaries—without aggregating subtree documents—could improve interpretability and efficiency, while entropy and category utility measures may enable autonomous selection of appropriate topic granularity.


\bibliography{custom}

@inproceedings{hldablei,
author = {Blei, David M. and Jordan, Michael I. and Griffiths, Thomas L. and Tenenbaum, Joshua B.},
title = {Hierarchical topic models and the nested chinese restaurant process},
year = {2003},
publisher = {MIT Press},
address = {Cambridge, MA, USA},
booktitle = {Proceedings of the 17th International Conference on Neural Information Processing Systems},
pages = {17–24},
numpages = {8},
location = {Whistler, British Columbia, Canada},
series = {NIPS'03}
}

@misc{pham2024topicgpt,
      title={TopicGPT: A Prompt-based Topic Modeling Framework}, 
      author={Chau Minh Pham and Alexander Hoyle and Simeng Sun and Philip Resnik and Mohit Iyyer},
      year={2024},
      eprint={2311.01449},
      archivePrefix={arXiv},
      primaryClass={cs.CL},
      url={https://arxiv.org/abs/2311.01449}, 
}

@misc{wu2024fastopic,
      title={FASTopic: Pretrained Transformer is a Fast, Adaptive, Stable, and Transferable Topic Model}, 
      author={Xiaobao Wu and Thong Nguyen and Delvin Ce Zhang and William Yang Wang and Anh Tuan Luu},
      year={2024},
      eprint={2405.17978},
      archivePrefix={arXiv},
      primaryClass={cs.CL},
      url={https://arxiv.org/abs/2405.17978}, 
}

@article{perotte2011hierarchically,
  title={Hierarchically supervised latent Dirichlet allocation},
  author={Perotte, Adler and Wood, Frank and Elhadad, Noemie and Bartlett, Nicholas},
  journal={Advances in neural information processing systems},
  volume={24},
  year={2011}
}

@inproceedings{mimno2007mixtures,
  title={Mixtures of hierarchical topics with pachinko allocation},
  author={Mimno, David and Li, Wei and McCallum, Andrew},
  booktitle={Proceedings of the 24th international conference on Machine learning},
  pages={633--640},
  year={2007}
}

@InProceedings{ARI-Topic-10.1007/978-3-662-44848-9_32,
author="Greene, Derek
and O'Callaghan, Derek
and Cunningham, P{\'a}draig",
editor="Calders, Toon
and Esposito, Floriana
and H{\"u}llermeier, Eyke
and Meo, Rosa",
title="How Many Topics? Stability Analysis for Topic Models",
booktitle="Machine Learning and Knowledge Discovery in Databases",
year="2014",
publisher="Springer Berlin Heidelberg",
address="Berlin, Heidelberg",
pages="498--513",
isbn="978-3-662-44848-9"
}

@article{barari2024incrementalconceptformationvisual,
      title={Incremental Concept Formation over Visual Images Without Catastrophic Forgetting}, 
      journal = {Advances in Cognitive Systems},
      author={Nicki Barari and Xin Lian and Christopher J. MacLellan},
      year={2024},
}

@inproceedings{coherence-10.1145/2684822.2685324,
author = {R\"{o}der, Michael and Both, Andreas and Hinneburg, Alexander},
title = {Exploring the Space of Topic Coherence Measures},
year = {2015},
isbn = {9781450333177},
publisher = {Association for Computing Machinery},
address = {New York, NY, USA},
url = {https://doi.org/10.1145/2684822.2685324},
doi = {10.1145/2684822.2685324},
booktitle = {Proceedings of the Eighth ACM International Conference on Web Search and Data Mining},
pages = {399–408},
numpages = {10},
location = {Shanghai, China},
series = {WSDM '15}
}

@INPROCEEDINGS{DBSTREAM-6906426,
  author={Bär, Arian and Casas, Pedro and Golab, Lukasz and Finamore, Alessandro},
  booktitle={2014 International Wireless Communications and Mobile Computing Conference (IWCMC)}, 
  title={DBStream: An online aggregation, filtering and processing system for network traffic monitoring}, 
  year={2014},
  volume={},
  number={},
  pages={611-616},
  doi={10.1109/IWCMC.2014.6906426}}

@article{efficientinductionlanguagemodels2022,
      title={Efficient Induction of Language Models Via Probabilistic Concept Formation},
      journal = {Advances in Cognitive Systems},
      author={Christopher J. MacLellan and Peter Matsakis and Pat Langley},
      year={2022},
}

@article{convolutionalcobweb2021,
      title={Convolutional Cobweb: A Model of Incremental Learning from 2D Images},
      journal = {Advances in Cognitive Systems},
      author={Christopher J. MacLellan and Harshil Thakur},
      year={2021},
}

@article{wang2025taxonomic,
  title={Taxonomic Networks: A Representation for Neuro-Symbolic Pairing},
  author={Wang, Zekun and Haarer, Ethan L and Barari, Nicki and MacLellan, Christopher J},
  journal={arXiv preprint arXiv:2505.24601},
  year={2025}
}

@article{barari2024avoiding,
  title={Avoiding catastrophic forgetting in visual classification using human concept formation},
  author={Barari, Nicki and Lian, Xin and MacLellan, Christopher J},
  journal={CoRR},
  year={2024}
}

@article{LIAN2025101371,
title = {Efficient and scalable masked word prediction using concept formation},
journal = {Cognitive Systems Research},
volume = {92},
pages = {101371},
year = {2025},
issn = {1389-0417},
doi = {https://doi.org/10.1016/j.cogsys.2025.101371},
url = {https://www.sciencedirect.com/science/article/pii/S1389041725000518},
author = {Xin Lian and Zekun Wang and Christopher J. MacLellan}
}

@inproceedings{gluck1985information,
  title={Information, uncertainty, and the utility of categories},
  author={Gluck, Mark A and Corter, James E},
  booktitle={Proceedings of the Annual Meeting of the Cognitive Science Society},
  volume={7},
  year={1985}
}

@article{rosch1975family,
  title        = {Family Resemblances: Studies in the Internal Structure of Categories},
  author       = {Rosch, Eleanor and Mervis, Carolyn B.},
  journal      = {Cognitive Psychology},
  volume       = {7},
  number       = {4},
  pages        = {573--605},
  year         = {1975},
  doi          = {10.1016/0010-0285(75)90024-9}
}

@article{mcinnes2018umap-software,
  title={UMAP: Uniform Manifold Approximation and Projection},
  author={McInnes, Leland and Healy, John and Saul, Nathaniel and Grossberger, Lukas},
  journal={The Journal of Open Source Software},
  volume={3},
  number={29},
  pages={861},
  year={2018}
}

@inproceedings{mini-batch-kmeans-10.1145/1772690.1772862,
author = {Sculley, D.},
title = {Web-scale k-means clustering},
year = {2010},
isbn = {9781605587998},
publisher = {Association for Computing Machinery},
address = {New York, NY, USA},
url = {https://doi.org/10.1145/1772690.1772862},
doi = {10.1145/1772690.1772862},
booktitle = {Proceedings of the 19th International Conference on World Wide Web},
pages = {1177–1178},
numpages = {2},
location = {Raleigh, North Carolina, USA},
series = {WWW '10}
}

@inproceedings{viegas2020cluhtm,
    title = "{C}lu{HTM} - Semantic Hierarchical Topic Modeling based on {C}lu{W}ords",
    author = "Viegas, Felipe  and
      Cunha, Washington  and
      Gomes, Christian  and
      Pereira, Ant{\^o}nio  and
      Rocha, Leonardo  and
      Goncalves, Marcos",
    editor = "Jurafsky, Dan  and
      Chai, Joyce  and
      Schluter, Natalie  and
      Tetreault, Joel",
    booktitle = "Proceedings of the 58th Annual Meeting of the Association for Computational Linguistics",
    month = jul,
    year = "2020",
    address = "Online",
    publisher = "Association for Computational Linguistics",
    url = "https://aclanthology.org/2020.acl-main.724/",
    doi = "10.18653/v1/2020.acl-main.724",
    pages = "8138--8150"
}

@misc{shahid2023hyhtm,
      title={HyHTM: Hyperbolic Geometry based Hierarchical Topic Models}, 
      author={Simra Shahid and Tanay Anand and Nikitha Srikanth and Sumit Bhatia and Balaji Krishnamurthy and Nikaash Puri},
      year={2023},
      eprint={2305.09258},
      archivePrefix={arXiv},
      primaryClass={cs.IR},
      url={https://arxiv.org/abs/2305.09258}, 
}

@article{kingma2013auto,
  title={Auto-encoding variational bayes},
  author={Kingma, Diederik P and Welling, Max},
  journal={arXiv preprint arXiv:1312.6114},
  year={2013}
}

@inproceedings{chen2021evaluating,
    title = "Tree-Structured Topic Modeling with Nonparametric Neural Variational Inference",
    author = "Chen, Ziye  and
      Ding, Cheng  and
      Zhang, Zusheng  and
      Rao, Yanghui  and
      Xie, Haoran",
    editor = "Zong, Chengqing  and
      Xia, Fei  and
      Li, Wenjie  and
      Navigli, Roberto",
    booktitle = "Proceedings of the 59th Annual Meeting of the Association for Computational Linguistics and the 11th International Joint Conference on Natural Language Processing (Volume 1: Long Papers)",
    month = aug,
    year = "2021",
    address = "Online",
    publisher = "Association for Computational Linguistics",
    url = "https://aclanthology.org/2021.acl-long.182/",
    doi = "10.18653/v1/2021.acl-long.182",
    pages = "2343--2353"
}

@article{lang1995newsweeder,
  title={NewsWeeder: Learning to filter netnews},
  author={Lang, Ken},
  journal={Proceedings of the Twelfth International Conference on Machine Learning (ICML)},
  year={1995}
}

@inproceedings{STACKOVERFLOW,
author = {Movshovitz-Attias, Dana and Movshovitz-Attias, Yair and Steenkiste, Peter and Faloutsos, Christos},
title = {Analysis of the reputation system and user contributions on a question answering website: StackOverflow},
year = {2013},
isbn = {9781450322409},
publisher = {Association for Computing Machinery},
address = {New York, NY, USA},
url = {https://doi.org/10.1145/2492517.2500242},
doi = {10.1145/2492517.2500242},
booktitle = {Proceedings of the 2013 IEEE/ACM International Conference on Advances in Social Networks Analysis and Mining},
pages = {886–893},
numpages = {8},
location = {Niagara, Ontario, Canada},
series = {ASONAM '13}
}

@misc{liu2019robertarobustlyoptimizedbert,
      title={RoBERTa: A Robustly Optimized BERT Pretraining Approach}, 
      author={Yinhan Liu and Myle Ott and Naman Goyal and Jingfei Du and Mandar Joshi and Danqi Chen and Omer Levy and Mike Lewis and Luke Zettlemoyer and Veselin Stoyanov},
      year={2019},
      eprint={1907.11692},
      archivePrefix={arXiv},
      primaryClass={cs.CL},
      url={https://arxiv.org/abs/1907.11692}, 
}

@Article{cu-corter-gluck-1992-18581-001,
author={Corter, James E.
and Gluck, Mark A.},
title={Explaining basic categories: Feature predictability and information.},
journal={Psychological Bulletin},
year={1992},
publisher={American Psychological Association},
address={US},
volume={111},
number={2},
pages={291-303},
doi={10.1037/0033-2909.111.2.291},
url={https://doi.org/10.1037/0033-2909.111.2.291}
}

@article{Wu_2024_survey,
   title={A survey on neural topic models: methods, applications, and challenges},
   volume={57},
   ISSN={1573-7462},
   url={http://dx.doi.org/10.1007/s10462-023-10661-7},
   DOI={10.1007/s10462-023-10661-7},
   number={2},
   journal={Artificial Intelligence Review},
   publisher={Springer Science and Business Media LLC},
   author={Wu, Xiaobao and Nguyen, Thong and Luu, Anh Tuan},
   year={2024},
   month=jan }

@misc{zheng2013supervisedneuralautoregressivetopic,
      title={A Supervised Neural Autoregressive Topic Model for Simultaneous Image Classification and Annotation}, 
      author={Yin Zheng and Yu-Jin Zhang and Hugo Larochelle},
      year={2013},
      eprint={1305.5306},
      archivePrefix={arXiv},
      primaryClass={cs.CV},
      url={https://arxiv.org/abs/1305.5306}, 
}

@article{scikitlearn,
  author       = {Fabian Pedregosa and
                  Ga{\"{e}}l Varoquaux and
                  Alexandre Gramfort and
                  Vincent Michel and
                  Bertrand Thirion and
                  Olivier Grisel and
                  Mathieu Blondel and
                  Peter Prettenhofer and
                  Ron Weiss and
                  Vincent Dubourg and
                  Jake VanderPlas and
                  Alexandre Passos and
                  David Cournapeau and
                  Matthieu Brucher and
                  Matthieu Perrot and
                  Edouard Duchesnay},
  title        = {Scikit-learn: Machine Learning in Python},
  journal      = {CoRR},
  volume       = {abs/1201.0490},
  year         = {2012},
  url          = {http://arxiv.org/abs/1201.0490},
  eprinttype    = {arXiv},
  eprint       = {1201.0490},
  bibsource    = {dblp computer science bibliography, https://dblp.org}
}

@article{zhang2015character,
  title={Character-level convolutional networks for text classification},
  author={Zhang, Xiang and Zhao, Junbo and LeCun, Yann},
  journal={Advances in Neural Information Processing Systems (NeurIPS)},
  year={2015}
}

@article{dieng2020topic,
    title = "Topic Modeling in Embedding Spaces",
    author = "Dieng, Adji B.  and
      Ruiz, Francisco J. R.  and
      Blei, David M.",
    editor = "Johnson, Mark  and
      Roark, Brian  and
      Nenkova, Ani",
    journal = "Transactions of the Association for Computational Linguistics",
    volume = "8",
    year = "2020",
    address = "Cambridge, MA",
    publisher = "MIT Press",
    url = "https://aclanthology.org/2020.tacl-1.29/",
    doi = "10.1162/tacl_a_00325",
    pages = "439--453"
}

@article{duan2021sawetm,
  author       = {Zhibin Duan and
                  Dongsheng Wang and
                  Bo Chen and
                  Chaojie Wang and
                  Wenchao Chen and
                  Yewen Li and
                  Jie Ren and
                  Mingyuan Zhou},
  title        = {Sawtooth Factorial Topic Embeddings Guided Gamma Belief Network},
  journal      = {CoRR},
  volume       = {abs/2107.02757},
  year         = {2021},
  url          = {https://arxiv.org/abs/2107.02757},
  eprinttype    = {arXiv},
  eprint       = {2107.02757},
  bibsource    = {dblp computer science bibliography, https://dblp.org}
}

@misc{gupta2025retrieval,
      title={Hierarchical Semantic Retrieval with Cobweb}, 
      author={Anant Gupta and Karthik Singaravadivelan and Zekun Wang},
      year={2025},
      eprint={2510.02539},
      archivePrefix={arXiv},
      primaryClass={cs.CL},
      url={https://arxiv.org/abs/2510.02539}, 
}

@inproceedings{wu2024traco,
    title        = {On the Affinity, Rationality, and Diversity of Hierarchical Topic Modeling},
    author       = {Wu, Xiaobao and Pan, Fengjun and Nguyen, Thong and Feng, Yichao and Liu, Chaoqun and Nguyen, Cong-Duy and Luu, Anh Tuan},
    year         = 2024,
    booktitle    = {Proceedings of the AAAI Conference on Artificial Intelligence},
    url          = {https://arxiv.org/pdf/2401.14113.pdf}
}

@article{lu2024self,
  title={Self-supervised Topic Taxonomy Discovery in the Box Embedding Space},
  author={Lu, Yuyin and Chen, Hegang and Mao, Pengbo and Rao, Yanghui and Xie, Haoran and Wang, Fu Lee and Li, Qing},
  journal={Transactions of the Association for Computational Linguistics},
  volume={12},
  pages={1401--1416},
  year={2024},
  publisher={MIT Press 255 Main Street, 9th Floor, Cambridge, Massachusetts 02142, USA~…}
}

@misc{koltcov2021analysistuninghierarchicaltopic,
      title={Analysis and tuning of hierarchical topic models based on Renyi entropy approach}, 
      author={Sergei Koltcov and Vera Ignatenko and Maxim Terpilovskii and Paolo Rosso},
      year={2021},
      eprint={2101.07598},
      archivePrefix={arXiv},
      primaryClass={stat.ML},
      url={https://arxiv.org/abs/2101.07598}, 
}

@inproceedings{isonuma2020tree,
    title = "{T}ree-{S}tructured {N}eural {T}opic {M}odel",
    author = "Isonuma, Masaru  and
      Mori, Junichiro  and
      Bollegala, Danushka  and
      Sakata, Ichiro",
    editor = "Jurafsky, Dan  and
      Chai, Joyce  and
      Schluter, Natalie  and
      Tetreault, Joel",
    booktitle = "Proceedings of the 58th Annual Meeting of the Association for Computational Linguistics",
    month = jul,
    year = "2020",
    address = "Online",
    publisher = "Association for Computational Linguistics",
    url = "https://aclanthology.org/2020.acl-main.73/",
    doi = "10.18653/v1/2020.acl-main.73",
    pages = "800--806"
}

@article{LDA-10.5555/944919.944937,
author = {Blei, David M. and Ng, Andrew Y. and Jordan, Michael I.},
title = {Latent dirichlet allocation},
year = {2003},
issue_date = {3/1/2003},
publisher = {JMLR.org},
volume = {3},
number = {null},
issn = {1532-4435},
journal = {J. Mach. Learn. Res.},
month = mar,
pages = {993–1022},
numpages = {30}
}

@inproceedings{lifelong-neuraltm-10.5555/3524938.3525304,
author = {Gupta, Pankaj and Chaudhary, Yatin and Runkler, Thomas and Sch\"{u}tze, Hinrich},
title = {Neural topic modeling with continual lifelong learning},
year = {2020},
publisher = {JMLR.org},
booktitle = {Proceedings of the 37th International Conference on Machine Learning},
articleno = {366},
numpages = {11},
series = {ICML'20}
}

@inproceedings{OnlineLDA-NIPS2010_71f6278d,
 author = {Hoffman, Matthew and Bach, Francis and Blei, David},
 booktitle = {Advances in Neural Information Processing Systems},
 editor = {J. Lafferty and C. Williams and J. Shawe-Taylor and R. Zemel and A. Culotta},
 pages = {},
 publisher = {Curran Associates, Inc.},
 title = {Online Learning for Latent Dirichlet Allocation},
 url = {https://proceedings.neurips.cc/paper_files/paper/2010/file/71f6278d140af599e06ad9bf1ba03cb0-Paper.pdf},
 volume = {23},
 year = {2010}
}

@article{fisher1987cobweb,
  title={Knowledge acquisition via incremental conceptual clustering},
  author={Fisher, Douglas H.},
  journal={Machine Learning},
  volume={2},
  number={2},
  pages={139--172},
  year={1987},
  publisher={Springer}
}

@misc{luo2025empiricalstudycatastrophicforgetting,
      title={An Empirical Study of Catastrophic Forgetting in Large Language Models During Continual Fine-tuning}, 
      author={Yun Luo and Zhen Yang and Fandong Meng and Yafu Li and Jie Zhou and Yue Zhang},
      year={2025},
      eprint={2308.08747},
      archivePrefix={arXiv},
      primaryClass={cs.CL},
      url={https://arxiv.org/abs/2308.08747}, 
}

@article{grootendorst2022bertopic,
  title={BERTopic: Neural topic modeling with a class-based TF-IDF procedure},
  author={Grootendorst, Maarten},
  journal={arXiv preprint arXiv:2203.05794},
  year={2022}
}

@article{Wu_2024,
   title={A survey on neural topic models: methods, applications, and challenges},
   volume={57},
   ISSN={1573-7462},
   url={http://dx.doi.org/10.1007/s10462-023-10661-7},
   DOI={10.1007/s10462-023-10661-7},
   number={2},
   journal={Artificial Intelligence Review},
   publisher={Springer Science and Business Media LLC},
   author={Wu, Xiaobao and Nguyen, Thong and Luu, Anh Tuan},
   year={2024},
   month=jan }

@misc{spatiotemporalDataset2025,
title={Space-Time MiniLM},
author={Haidar Jomaa},
year={2025},
url={https://huggingface.co/HaidarJomaa/Space-Time-MiniLM-v0}
}

@inproceedings{ushio-etal-2022-tweet,
    title = "{N}amed {E}ntity {R}ecognition in {T}witter: {A} {D}ataset and {A}nalysis on {S}hort-{T}erm {T}emporal {S}hifts",
    author = "Ushio, Asahi  and
        Neves, Leonardo  and
        Silva, Vitor  and
        Barbieri, Francesco. and
        Camacho-Collados, Jose",
    booktitle = "The 2nd Conference of the Asia-Pacific Chapter of the Association for Computational Linguistics and the 12th International Joint Conference on Natural Language Processing",
    month = nov,
    year = "2022",
    address = "Online",
    publisher = "Association for Computational Linguistics",
}

\appendix

\section{Risks}
\label{app:risks}

Like all topic models, \textsc{CobwebTM} inherits and can amplify biases present in its training data, such as over-representing dominant viewpoints while marginalizing minority perspectives or sensitive topics. The hierarchical structure may further legitimize biased or stereotypical groupings by presenting them as coherent topics, which can mislead downstream analysis or decision-making. Additionally, because topic models abstract language into latent structures without grounding or normative judgment, they risk obscuring harmful associations in data and being misused to draw causal or normative conclusions from biased text corpora.

\section{Datasets and Preprocessing}
\subsection{Datasets}
\label{app:datasets}
We use a diverse collection of datasets spanning news media, online forums, and social media to evaluate our models. This appendix provides a brief description of each dataset and the corresponding splits used in our experiments.

\paragraph{Spatiotemporal News Dataset}
We use the Spatiotemporal News Dataset~\citep{spatiotemporalDataset2025}, which consists of approximately 1.2 million English-language news articles collected from major news outlets across North America and annotated with temporal and geographic metadata. For our experiments, we randomly sample 5k documents from the test split.

\paragraph{Stack Overflow Dataset.}
We use the Stack Overflow Dataset~\citep{STACKOVERFLOW}, which contains question-and-answer forum posts covering a broad range of technical topics in computer science and software engineering. We randomly sample 5k posts spanning diverse subject areas.

\paragraph{TweetNER7 Dataset.}
We use the TweetNER7 Dataset~\citep{ushio-etal-2022-tweet}, which consists of short-form posts from X.com (formerly Twitter) across a variety of topics. We use the provided test split, which contains approximately 2.8k tweets.

\paragraph{20 Newsgroups.}
The 20 Newsgroups dataset~\citep{lang1995newsweeder} contains 18,846 documents evenly distributed across 20 discussion groups and is a standard benchmark for topic modeling and text clustering. We use the version distributed through the \texttt{scikit-learn} library~\citep{scikitlearn}.

\paragraph{AG News.}
We use the AG’s News Topic Classification Dataset (AG News)~\citep{zhang2015character}, which consists of news articles collected from Associated Press and Google News sources and categorized into four high-level topic classes. We randomly sample 50k documents from the training set and use the full test set consisting of approximately 7.6k documents.

All datasets used in our experiments are publicly available on the Hugging Face Hub and licensed for research use. While curated with privacy in mind, some datasets (e.g., Stack Overflow or TweetNER7) may contain identifiable or offensive content. We rely on the maintainers’ preprocessing and licensing terms and do not perform additional filtering. No extra personally identifying information is collected, stored, or exposed.

\subsection{Dataset Preprocessing}

\label{app:preprocessing}
To preprocess the datasets, we follow the steps in \cite{wu2024traco}: (1) tokenize documents and convert them to lowercase; (2) remove numbers, punctuations, and stopwords; (3) remove tokens with less than 3 characters.

\begin{table*}[ht]
\centering
\begin{tabular}{|l|cccc|}
\hline
Dataset & \multicolumn{4}{c|}{StackExchange} \\
\cline{2-5}
Model & None & 16 & 128 & 512 \\
\hline
COBWEBTM (ours) & \textbf{0.6131} & \textbf{0.6331} & \textbf{0.6227} & \textbf{0.6383} \\
BERTopic (MiniBatchKMeans) & 0.4234 & 0.4377 & 0.4202 & 0.4393 \\
BERTopic Refit (HDBSCAN) & 0.4249 & 0.4829 & 0.4666 & 0.4713 \\
BERTopic Refit (KMeans) & \underline{0.5507} & 0.5068 & 0.5110 & 0.5123 \\
Lifelong NTM & 0.1840 & 0.1621 & 0.1868 & 0.2057 \\
BERTopic (DBSTREAM) & 0.4569 & 0.4309 & 0.3872 & 0.4247 \\
Online LDA & 0.5199 & \underline{0.5663} & \underline{0.5332} & \underline{0.5479} \\
\hline
\end{tabular}
\caption{Final C$_v$ by UMAP dimensionality on StackExchange.}
\label{tab:umap_stackexchange}
\end{table*}

\begin{table*}[ht]
\centering
\begin{tabular}{|l|cccc|}
\hline
Dataset & \multicolumn{4}{c|}{TweetNER} \\
\cline{2-5}
Model & None & 16 & 128 & 512 \\
\hline
COBWEBTM (ours) & \textbf{0.7414} & \textbf{0.7249} & \textbf{0.7581} & \textbf{0.7048} \\
BERTopic (MiniBatchKMeans) & 0.4129 & 0.3531 & 0.3987 & 0.3807 \\
BERTopic Refit (HDBSCAN) & \underline{0.6733} & \underline{0.6501} & \underline{0.6524} & 0.4558 \\
BERTopic Refit (KMeans) & 0.3694 & 0.4131 & 0.4094 & 0.4111 \\
Lifelong NTM & 0.2811 & 0.3042 & 0.3446 & 0.1684 \\
BERTopic (DBSTREAM) & 0.6019 & 0.3951 & 0.4655 & \underline{0.5373} \\
Online LDA & 0.3171 & 0.3171 & 0.3171 & 0.3171 \\
\hline
\end{tabular}
\caption{Final C$_v$ by UMAP dimensionality on TweetNER.}
\label{tab:umap_tweetner}
\end{table*}

\begin{table*}[ht]
\centering
\begin{tabular}{|l|cccc|}
\hline
Dataset & \multicolumn{4}{c|}{Spatiotemporal} \\
\cline{2-5}
Model & None & 16 & 128 & 512 \\
\hline
COBWEBTM (ours) & \textbf{0.7960} & \textbf{0.6948} & \textbf{0.6837} & 0.6284 \\
BERTopic (MiniBatchKMeans) & \underline{0.6460} & \underline{0.6068} & \underline{0.6403} & \underline{0.6322} \\
BERTopic Refit (HDBSCAN) & 0.6572 & 0.5103 & 0.5190 & 0.5105 \\
BERTopic Refit (KMeans) & 0.6136 & 0.5568 & 0.5471 & 0.5516 \\
Lifelong NTM & 0.2381 & 0.1211 & 0.2331 & 0.1191 \\
BERTopic (DBSTREAM) & 0.4176 & 0.4109 & 0.4416 & 0.3964 \\
Online LDA & 0.4221 & 0.4163 & 0.4334 & 0.4185 \\
\hline
\end{tabular}
\caption{Final C$_v$ by UMAP dimensionality on Spatiotemporal dataset.}
\label{tab:umap_spatiotemporal}
\end{table*}

\section{Ablation Studies}
\label{app:ablations}

\subsection{Encoder Ablations}

We conduct ablation studies on the sentence encoder used to generate document embeddings. Our main experiments use \texttt{all-roberta-large-v1}, a RoBERTa Large model finetuned for information retrieval. To evaluate sensitivity to the embedding backbone, we also test two smaller encoders from the same family: \texttt{all-MiniLM-L12-v2} and \texttt{all-MiniLM-L6-v2}. Using models with similar training objectives but different parameter sizes allows us to isolate the effect of model capacity while controlling for architectural differences.

\begin{figure*}[ht]
        \includegraphics[width=\textwidth]{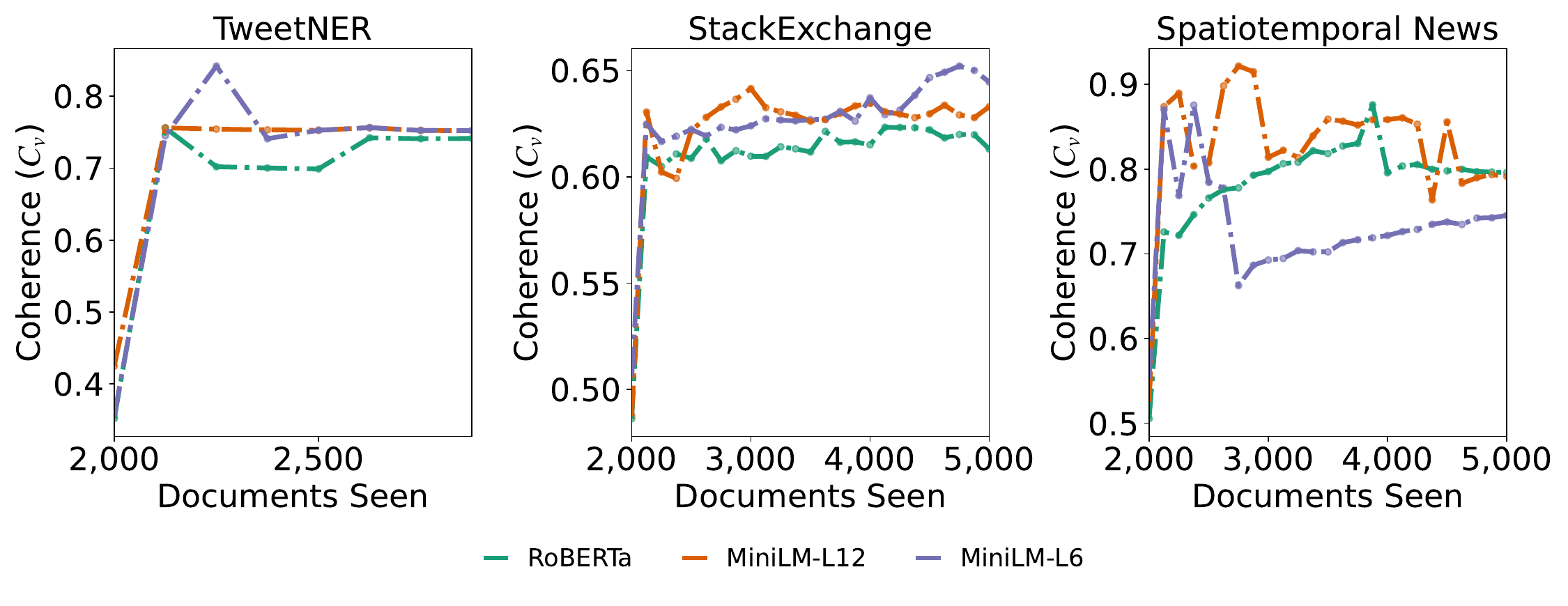}
    \caption{An ablation study to compare the lifelong results of using different embedding models for \textsc{CobwebTM}.}
    \label{fig:lifelong-encoder-ablation-results}
\end{figure*}
\begin{table*}[ht]
\centering
\begin{tabular}{|l|ccc|ccc|ccc|}
\hline
Dataset & \multicolumn{3}{c|}{20 News Groups} & \multicolumn{3}{c|}{AG News} & \multicolumn{3}{c|}{Stack Exchange} \\
\cline{2-4}\cline{5-7}\cline{8-10}
Model & NPMI & PCC & SD & NPMI & PCC & SD & NPMI & PCC & SD \\
\hline
MiniLM-L6 & \underline{0.205} & \underline{0.130} & 0.935 & 0.104 & 0.023 & 0.924 & 0.127 & \underline{0.070} & \underline{0.952}\\
MiniLM-L12 & 0.196 & 0.119 & \underline{0.952} & \textbf{0.109} & \textbf{0.031} & \textbf{0.959} & \underline{0.130} & {0.055} & 0.862 \\
RoBERTa (ours)   & \textbf{0.206} & \textbf{0.141} & \textbf{0.958} & \underline{0.108} & \underline{0.027} & \underline{0.942} & \textbf{0.131} & \textbf{0.073} & \textbf{0.959} \\
\hline
\end{tabular}
\caption{An ablation study to compare the hierarchical results of using different embedding models for \textsc{CobwebTM}.}
\label{tab:ablation_hierarchical_results}
\end{table*}

Table~\ref{tab:ablation_hierarchical_results} and Figure~\ref{fig:lifelong-encoder-ablation-results} show that \textsc{CobwebTM} is robust to encoder choice. \texttt{all-roberta-large-v1} achieves the strongest or near-strongest performance across most datasets, particularly on 20 Newsgroups and Stack Exchange. The MiniLM variants remain competitive despite their smaller size: \texttt{MiniLM-L6} performs comparably on 20 Newsgroups, while \texttt{MiniLM-L12} yields the best results on AG News. Overall, these results indicate that \textsc{CobwebTM} maintains stable hierarchical structure and lifelong topic quality even with lightweight embedding models, with larger encoders providing modest but consistent gains. One important observation, consistent with prior work on Cobweb for retrieval~\citep{gupta2025retrieval}, is that we restrict our study to encoders trained with distance-based objectives. This design choice is motivated by the Cobweb likelihood formulation, which relies on mean squared distance to compute log probabilities. As a result, encoders optimized for embedding-space distance alignment are better suited to the model’s underlying assumptions.

\subsection{Batch Size Ablations}

We also perform ablation studies with a reduced initial batch size for our lifelong topic modeling experiments. While real-world settings often contain a large corpus to pretrain a topic model with, instantiating it with stable topic definitions, there are many situations where a topic model has to be governed completely from scratch, necessitating smaller initial batch sizes.

\begin{figure*}[ht]
    \centering
    \begin{subfigure}[t]{0.3\textwidth}
        \centering
        \includegraphics[width=\textwidth]{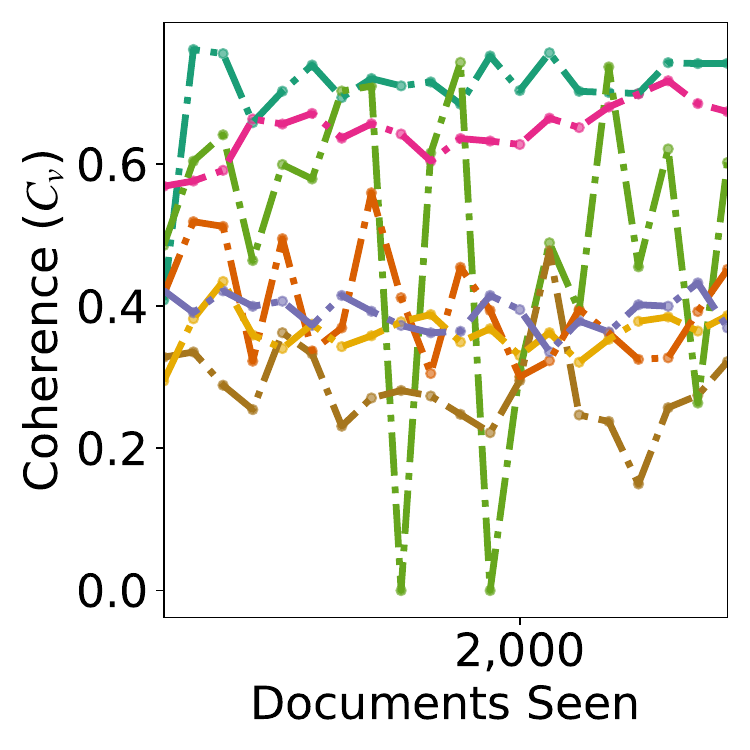}
        \caption{Tweetner Ablation}
        \label{fig:batch-size-tweetner-ablation}
    \end{subfigure}
    \hfill 
    \begin{subfigure}[t]{0.3\textwidth}
        \centering
        \includegraphics[width=\textwidth]{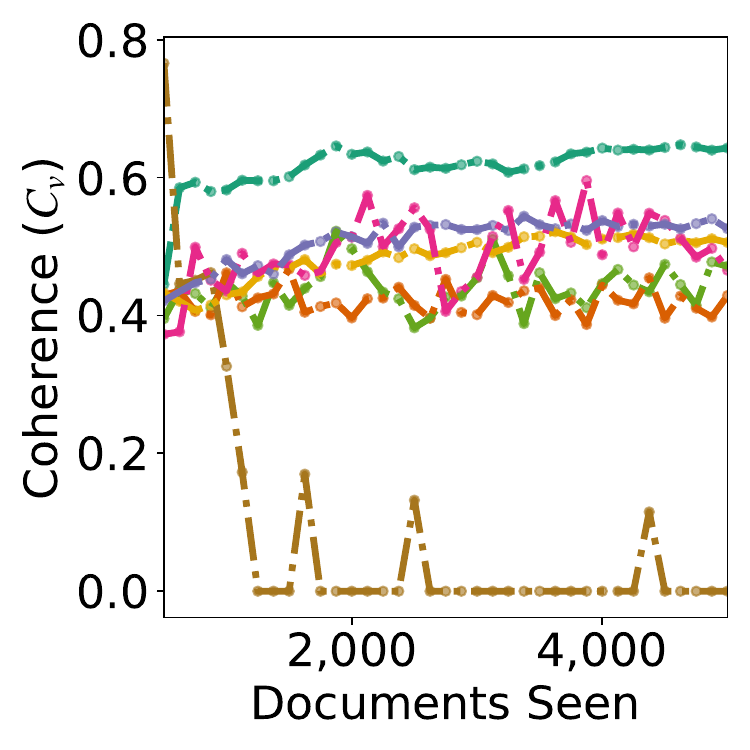}
        \caption{Stack Overflow Ablation}
        \label{fig:batch-size-stackexchange-ablation}
    \end{subfigure}
    \hfill 
    \begin{subfigure}[t]{0.3\textwidth}
        \centering
        \includegraphics[width=\textwidth]{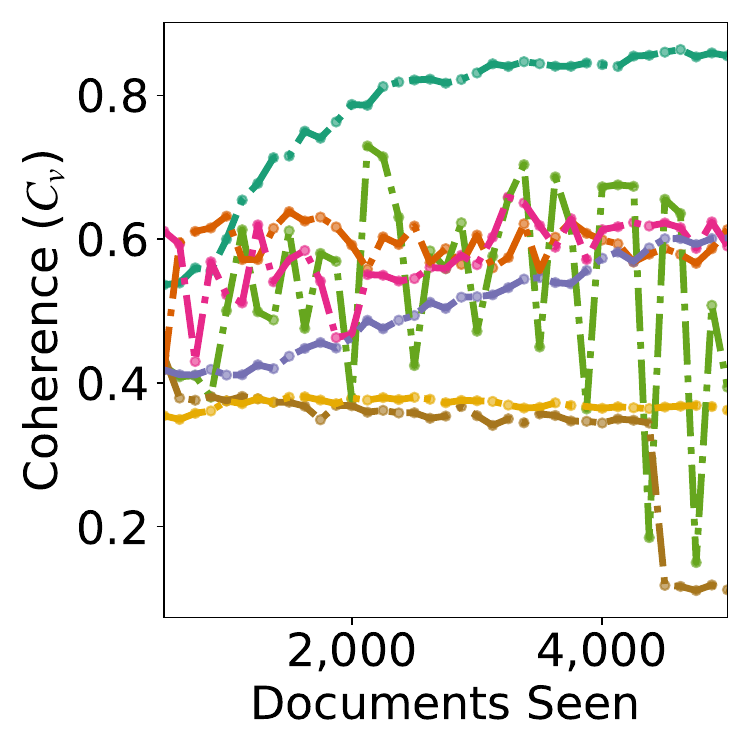}
        \caption{Spatiotemporal News Ablation}
        \label{fig:batch-size-spatiotemporal-ablation}
    \end{subfigure}
    \caption{An ablation study to compare the lifelong results of \textsc{CobwebTM} to other methods across all datasets with a reduced initial batch size of 500 documents.}
    \label{fig:appendix-batchsize-ablation-results}
\end{figure*}

We show the results of reducing the initial batch size to 500 on all three datasets in Figure \ref{fig:appendix-batchsize-ablation-results}. We see that \textsc{CobwebTM} still comfortably outperforms state-of-the-art lifelong methods, owing to its fundamental design around piecemeal learning, and is able to balance constructing new topics with maintaining old topics.

\subsection{UMAP Ablations}

We perform ablation studies to investigate the effects of dimensionality reduction on our framework with respect to other topic modeling frameworks. Specifically, we investigate the effects of UMAP \cite{mcinnes2018umap-software} on the RoBERTA embeddings used in CobwebTM and all BERTopic pipelines. We compare no UMAP applied to UMAP with $\text{d}=16, 128, 512$. As shown in figures \ref{tab:umap_spatiotemporal}, \ref{tab:umap_stackexchange} and \ref{tab:umap_tweetner}, Cobweb still performs best across all levels of UMAP, but there is no statistically significant difference in the results based on the level of UMAP applied, which is why we chose to omit it from the main paper.

\subsection{Incremental Ablations}

\subsubsection{Holdout Ablation}

To analyze the efficiency of incremental solutions in being able to create new topics to support, we ran an ablation with the TweetNER dataset, holding out the \textit{person} category and streaming the rest of the documents as per the default training setup, with the last batch consisting exclusively of the \textit{person} category. As shown in Figure \ref{tab:holdout_tweetner}, \textsc{CobwebTM}, like the Refit pipelines, is able to correctly create a document for the new category, while the other incremental solutions fail.s

\begin{table*}[ht]
\centering
\begin{tabular}{|l|cc|}
\hline
Dataset & \multicolumn{2}{c|}{TweetNER (Holdout: \textit{person})} \\
\cline{2-3}
Model & Before Injection C$_v$ & After Injection C$_v$ \\
\hline
COBWEBTM (ours) & \textbf{0.6425} & \textbf{1.0000} \\
BERTopic (MiniBatchKMeans) & 0.3812 & 0.0000 \\
BERTopic Refit (HDBSCAN) & 0.4845 & \textbf{1.0000} \\
BERTopic Refit (KMeans) & 0.3940 & \textbf{1.0000} \\
Lifelong NTM & 0.6120  & 0.0000 \\
BERTopic (DBSTREAM) & 0.3018 & 0.0000 \\
Online LDA & 0.3357 & 0.8132 \\
\hline
\end{tabular}
\caption{Holdout topic injection experiment on TweetNER dataset.}
\label{tab:holdout_tweetner}
\end{table*}

\subsubsection{Cobweb Operation Analysis}

\begin{figure*}[ht]
    \includegraphics[width=\textwidth]{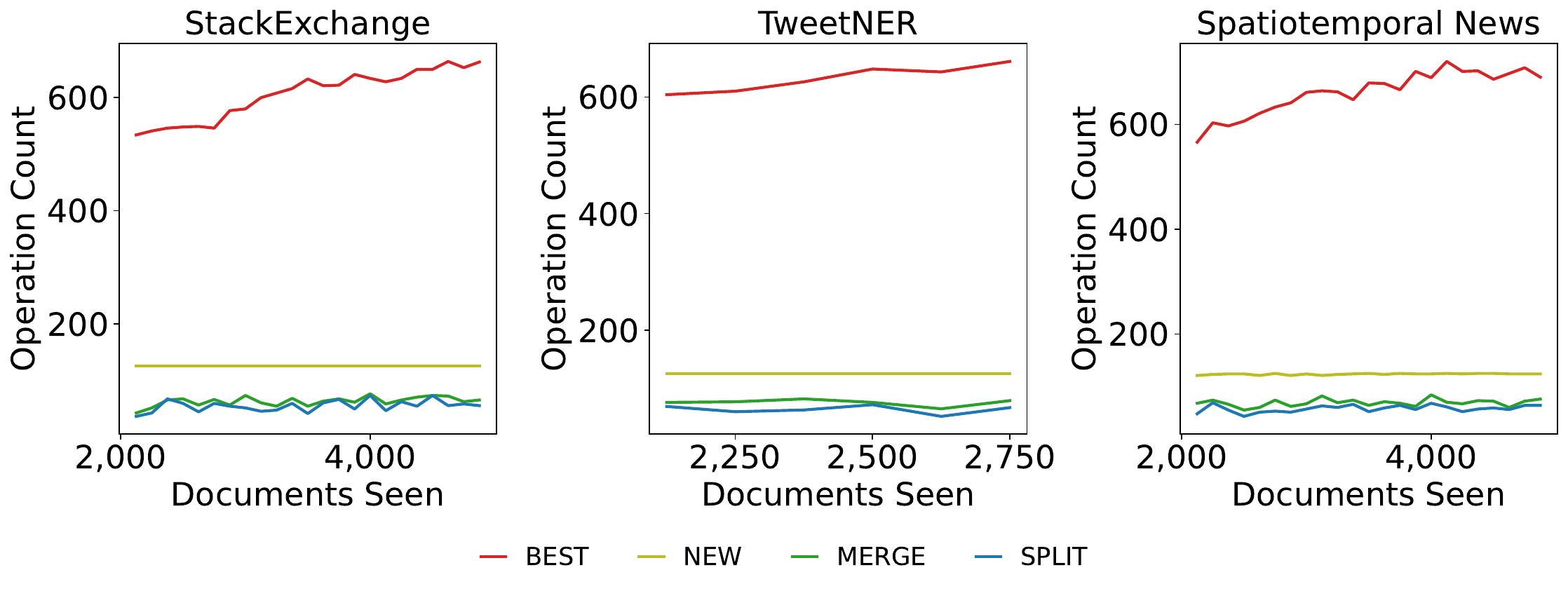}
\caption{An ablation study to compare the amount of each Cobweb operation per-batch for all three datasets in \textsc{CobwebTM}.}
\label{fig:incremental-cobweb-ops-ablation}
\end{figure*}

We conducted an analysis of the frequency of the different node operations in Cobweb with each dataset, shown in figure \ref{fig:incremental-cobweb-ops-ablation}. The most common operation is the "INSERT" operation, necessary for traversing the tree as a whole. Notably, the "INSERT" operation occurs at a log-likeAlso importantly, the MERGE and SPLIT operations each happen ~10\% of the time across batches, showcasing our method's ability to restructure clusters and create new clusters only when necessary. In this way, COBWEBTM is able to adapt its level of restructuring with respect to the size of its tree, a more robust solution than a fixed number of MERGEs and SPLITs. Additionally, NEW is only used to create a new leaf node for each document, which is why it happens exactly 125 times for each batch of 125 documents.

\subsubsection{Low-Batch-Size Ablation}

We ran an additional experiment with 5,000 documents from the StackExchange Dataset comparing our model's runtime to baselines, measuring the time taken to update topics for batches of size 10, with results shown in . The Cobweb algorithm does not have a batch-size parameter, training on one instance at a time, so at smaller batch-sizes, our model performs robustly while maintaining the best performance. Due to Cobweb’s heuristic-guided best first search, it explores approximately $\log(n)$ nodes per update, making it robust as the size of the tree increases.

\begin{table*}[ht]
\centering
\begin{tabular}{|l|cc|}
\hline
Dataset & \multicolumn{2}{c|}{StackExchange (batch\_size=10)} \\
\cline{2-3}
Model & Total Time (s) & Final Batch (s) \\
\hline
COBWEBTM (ours) & 382.94 & 1.37 \\
BERTopic (MiniBatchKMeans) & 362.81 & 1.52 \\
BERTopic Refit (HDBSCAN) & 16525.93 & 87.79 \\
BERTopic Refit (KMeans) & 10749.74 & 50.93 \\
Lifelong NTM & 657.44 & 2.62 \\
BERTopic (DBSTREAM) & 805.71 & 3.62 \\
Online LDA & \textbf{25.17} & \textbf{0.07} \\
\hline
\end{tabular}
\caption{Runtime comparison on StackExchange dataset.}
\label{tab:runtime_stackexchange}
\end{table*}

\subsubsection{Corpus Size Ablations}

We ran additional experiments measuring our model's performance with 20,000 documents and 50,000 documents from the StackExchange Dataset, which we report in Figures \ref{tab:coherence_stackexchange_long_combined}. As each new instance begins at the root node of the Cobweb tree and populates as a leaf through a single, greedily chosen path, COBWEBTM updates an average of O(log(N)) nodes with an insertion of each new document, minimizing our total time complexity and allowing us to retain robustness over long document streams. The results of the Refit Pipelines are not shown as they timed out after 15 hours of training.

\begin{table*}[ht]
\centering
\begin{tabular}{|l|cc|cc|}
\hline
Dataset & \multicolumn{2}{c|}{20k StackExchange} & \multicolumn{2}{c|}{50k StackExchange} \\
\cline{2-3}\cline{4-5}
Model & Start C$_v$ & Final C$_v$ & Start C$_v$ & Final C$_v$ \\
\hline
COBWEBTM (ours) & 0.4932 & \underline{0.6573} & 0.5010 & \underline{0.6441} \\
BERTopic (MiniBatchKMeans) & 0.4810 & 0.4270 & 0.4958 & 0.4476 \\
BERTopic Refit (HDBSCAN) &  &  &  &  \\
BERTopic Refit (KMeans) &  &  &  &  \\
Lifelong NTM & \textbf{0.7133} & 0.5522 & \textbf{0.7752} & 0.0000 \\
BERTopic (DBSTREAM) & 0.4481 & 0.4722 & 0.4801 & 0.4144 \\
Online LDA & 0.4365 & \textbf{0.6474} & 0.4782 & \textbf{0.5909} \\
\hline
\end{tabular}
\caption{Coherence C$_v$ comparison on 20k and 50k StackExchange datasets.}
\label{tab:coherence_stackexchange_long_combined}
\end{table*}

\section{Use of Large Language Models}
Large language models (LLMs) were used for grammar refinement, integration of new material, and as coding assistants for structuring and implementation support. All technical content and ideas were developed by the authors, and any LLM-generated output was subsequently modified and verified by the authors.

\end{document}